# Semi-supervised Learning for Discrete Choice Models

Jie Yang[1], Sergey Shebalov[2], Diego Klabjan[3]

[1]Department of Civil and Environmental Engineering, Northwestern University, Evanston, IL, USA

[2]Sabre Holdings, Sabre Airline Solutions, Southlake, Texas

[3]Department of Industrial Engineering and Management Sciences, Northwestern University, Evanston, IL, USA

## Abstract

We introduce a semi-supervised discrete choice model to calibrate *discrete choice models* when relatively few requests have both choice sets and stated preferences but the majority only have the choice sets. Two classic *semi-supervised learning* algorithms, the *expectation maximization* algorithm and the *cluster-and-label* algorithm, have been adapted to our choice modeling problem setting. We also develop two new algorithms based on the *cluster-and-label* algorithm. The new algorithms use the *Bayesian Information Criterion* to evaluate a clustering setting to automatically adjust the number of clusters. Two computational studies including a hotel booking case and a large-scale airline itinerary shopping case are presented to evaluate the prediction accuracy and computational effort of the proposed algorithms. Algorithmic recommendations are rendered under various scenarios.

## 1. Introduction

Airlines and hotels are trying to encourage travelers to book service directly through their own channels. For example, Lufthansa is looking to make a stand by charging an extra $18 for every ticket issued via a *global distribution system* (GDS)—the technology behind the booking systems used by travel agents and online travel agencies (Economist, 2015). GDS providers whose biggest asset is the data deluge are facing more intense competition. Confronting the industry challenges, GDS solutions have to understand better travelers' behavior and preference, to predict travel demand and market attractiveness, and subsequently monetize such findings, and thus prevent a corner overtaking.

The travel industry creates lots of data. Yet, traditional travel demand forecasting studies rely either on collecting travelers' responses from designed survey experiments, or on reconstituting choice sets by adding inferred choice alternatives to stated preferences. For example, Carrier (2008) combined observed flight booking data with fare rules and seat availability data to reconstitute the choice set of each booking. Both approaches, however have their deficiencies. Conducting survey experiments is time consuming and labor intensive. It is hard to guarantee that the respondents have the actual travel demand, and their decision making process in a survey may be inconsistent with that in a real booking environment. In addition, the designed choice sets, in some cases (e.g. flight booking, hotel booking, etc.) are more restricted comparing to a real environment. Reconstituting choice sets is better but limited to simplified rules and assumptions. In summary, both approaches cannot reflect reality. Nowadays, storing large-scale travel requests (either for air travel or hotel) and returned services (e.g. itineraries a traveler see when booking through a travel agency) is possible by GDS providers. The actual bookings are captured by separate reservation systems that are not integrated with request and service information systems. In order to develop a discrete choice model (DCM), the actual stated preferences are needed. We employ a semi-supervised approach to derive the stated preferences. We first assume that if there is an itinerary in the choice set dominating all other itineraries with respect to the fare and deviation from the request parameters such as the departure time and the elapsed time, then such an itinerary would be the preferred choice. This strategy creates requests with preferred choices. For all remaining requests we employ semi-supervised techniques to infer the preferred choices. We refer to the unmatched requests as unlabeled data and the matched requests with a responding itinerary as labeled data. Previous studies concentrate on choice modeling with only labeled data but simply utilizing limited labeled data may lead to bias. Leaving out unlabeled data is wasteful as the unlabeled data also captures travelers' preferences (e.g. to book an itinerary, travelers are often required to state their preferred airline, cabin, departure and arrival times, etc.). In some situations, unlabeled data could offer information to segment travelers and potentially prevent bias. In order to leverage the value of unlabeled data, we consider semi-supervised learning (SSL) which lies between *supervised* and *unsupervised learning*. In *supervised learning*, there is a known, fixed set of categories and category-labeled training data used to induce a classification function. In contrast, *unsupervised learning* applies to unlabeled training data. Both *supervised* and *unsupervised learning* have been widely applied to transportation problems. For example, Zhang and Xie (2008) applied support vector machine which is a supervised learning method to travel mode choice modeling. Vlahogianni et al. (2008) developed a multilayer strategy that integrated the k-means algorithm to cluster traffic patterns. To the best of our knowledge, an SSL framework has not yet been applied in choice modeling.

In machine learning, the SSL algorithms have been widely used to improve prediction accuracy in classification problems with also unlabeled data. The SSL algorithms usually make use of the smoothness, cluster, and manifold assumptions, and can be roughly categorized into five categories: (1) self-training; (2) SSL with generative models; (3) semi-supervised support vector machines ($S^3VM$), or transductive $SVM$; (4) SSL with graphs; (5) SSL with committees (Zhu, 2008; Peng et al., 2015). We refer to Zhu (2008) and Chapelle et al. (2006) for details about SSL algorithms.

Generative models make the assumption that both labeled and unlabeled requests come from the same parametric model (Schwenker and Trentin, 2014). The SSL-DCM problem can be considered as a generative model problem and could be solved by traditional algorithm frameworks such as the *expectation maximization* (EM) algorithm which was first presented by Dempster et al. (1977) who brought together and formalized many of the commonalities of previously suggested iterative techniques for likelihood maximization with missing data. The *cluster-and-label* (CL) algorithm which is a heuristic model finds the label using labeled data within each cluster, and assigns labels to unlabeled requests within each cluster. Demiriz et al. (1999) related an unsupervised clustering method, labeled each cluster with class membership, and simultaneously optimized the misclassification error of the resulting clusters.

Besides the classic SSL algorithms such as the EM framework and CL applied to SSL-DCM, we introduce two new methods *X-cluster-and-label-1* (XCL1) and *X-cluster-and-label-2* (XCL2) based on the X-means algorithm (Pelleg and Moore, 2000). These algorithms have never been applied in the context of SSL and they have been designed to cope with the problem of setting the hyper parameter for the number of clusters required by CL. X-means is a method which includes model estimation inside the cluster partitioning procedure. It can efficiently search the space of cluster locations and number of clusters to optimize the *Bayesian Information Criterion* (BIC) or *Akaike Information Criterion* (AIC). Pelleg and Moore (2000) discovered that this technique could reveal the true number of classes in the underlying distribution and provide a fast, statistically founded estimation. In the XCL1 and XCL2 algorithms, we search partitions based on the entire data set and select the best result which optimizes the BIC of the labeled data set. An adapted BIC formula is derived to consider the maximized log-likelihood, the number of features and the number of requests.

As mentioned earlier, it is often difficult for transportation planners to collect a solid data set. Hence, discrete choice models with incomplete data have recently raised interest in transportation problems. Vulcano et al. (2010) developed a maximum likelihood estimation algorithm to use a variation of the EM method accounting for unobservable data with no-purchase outcomes in airline revenue management. Newman et al. (2012) applied the EM method to estimate the market share of one alternative that was not observed to be chosen in the estimation data set. Their case study aimed at demonstrating the consistency and potential viability of the methodology; significance levels of coefficient estimates are not reported and a more thorough evaluation of this needs to be conducted (Newman et al., 2012). Another type of incomplete data is called "choice-restricted" where there are unobserved choices in the choice set. Lindsey et al. (2013) only observed acceptance decisions but no alternatives. They used a supplementary sample (an unbiased sample with only features) to augment the likelihood function. All these works are different from our work. Vulcano et al. (2010) and Newman et al. (2012) solved the problem which has to only infer no-purchase alternatives. Lindsey et al. (2013) solved the problem when the data set only comes with the chosen choice. But in our problem, we do not infer the no-purchase alternatives and we have both chosen and unchosen choices in the labeled data. In addition, we have a rich set of unlabeled data.

In the computational experiments, we first apply a semi-supervised *ranked-ordered logit* (ROL) model on a hotel booking case to explore the capability and accuracy of the SSL-DCM algorithms developed herein (adapted EM, CL, XCL1 and XCL2). The hotel case is completely labeled and we use it to evaluate the predictive power of our models. The ROL model was introduced in the literature by Beggs et al. (1981) and has also been used in transportation studies (Podgorski and Kockelman, 2006; Calfee et al., 2001). The ROL model can be transformed into a series of *multinomial logit* (MNL) models: an MNL model for the most preferred item; another MNL model for the second-ranked item to be preferred over all items except the one ranked first, and so on (Fok et al., 2012). To evaluate the prediction accuracy, five metrics based on log-likelihood, Kendall's $\tau$, position difference and reciprocal rank difference are used. In the second case which herein motivated the entire work, we apply the SSL-DCM algorithms to an airline itinerary shopping data set.

Our contributions are summarized as follows:
1. We mathematically define the semi-supervised discrete choice problem. To the best of our knowledge, no prior studies have been focused on this topic.
2. We adapt two classic SSL algorithms to the semi-supervised discrete choice problem.
3. We design two new algorithms (not previously known in the SSL context) which include model estimation inside the partitioning procedures.
4. For the two case studies, we design different metrics, show that the proposed algorithms have a good prediction accuracy and also provide recommendations of applying the algorithms.

The rest of the paper is organized as follow. Section 2 formally states the problem and the algorithms. Section 3 presents two case studies.

## 2. Methodologies

2.1 Problem setting

In the SSL-DCM problem, the input consists of a labeled data set $D^L$ and an unlabeled data set $D^U$:

$$D^L = \{(X_1, L_1), (X_2, L_2), \dots, (X_n, L_n)\}$$
$$D^U = \{(X_{n+1}, L_{n+1}), (X_{n+2}, L_{n+2}), \dots, (X_{n+m}, L_{n+m})\},$$

where we assume that vectors $L_{n+1} = \cdots = L_{n+m} = \mathbf{0}$ and $eL_1 = \cdots = eL_m = 1$ (Here $e = (1, \dots 1)$.) The latter binary encodes the only selection in the choice set.

Let $D^{U^*} = \{(X_{n+1}, L^*_{n+1}), (X_{n+2}, L^*_{n+2}), \dots, (X_{n+m}, L^*_{n+m})\}$ indicate that $D^U$ has been assigned "soft-labels." We denote $I = \{1, 2, \dots, (n+m)\}$. We also have labeled requests $X^L$ and unlabeled requests $X^U$:

$$X^L = \{X_1, X_2, \dots, X_n\}$$
$$X^U = \{X_{n+1}, X_{n+2}, \dots, X_{n+m}\}.$$

Given set $x$, denote by $|x|$ the cardinality of $x$. In the traditional SSL problem, we have $X_i$ defined as a feature vector and $L_i \in \{1, 2, \dots, M\}$ where $L_i$ is a class variable. In the SSL-DCM problem, we define the choice set $X_i = \{s_{i1}, s_{i2}, \dots, s_{i|X_i|}\}$ where $s_{ij} \in \mathbb{R}^d$ is a feature vector and $|X_i|$ is the number of choices in $X_i$, and $L_i = \{\delta_{i1}, \delta_{i2}, \dots, \delta_{i|X_i|}\}$ as a set of binary indicators encoding selection. We also assume a feature split $s_{ij} = (s_{ij}^{ISF}, s_{ij}^{MSF})$ exists where $s_{ij}^{ISF}$ is a vector of individual or request specific features (e.g. respondent's income, gender and age group, etc.), and $s_{ij}^{MSF}$ is a vector of alternative specific features and/or interactive features (e.g. one itinerary's departure time, respondent's income interacting with itinerary's cabin class). So we have $s_{ij}^{ISF} =$ $v_i$ for every $j = 1, 2, \ldots, |X_i|$. Later we use $V = \{v_1, v_2, \ldots, v_{n+m}\}$. In $\{L_1, L_2, \ldots, L_n\}$ and $\{L^*_{n+1}, L^*_{n+2}, \ldots, L^*_{n+m}\}$, we have $\sum_{j=1}^{|X_i|} \delta_{ij} = 1$ while in $\{L_{n+1}, L_{n+2}, \ldots, L_{n+m}\}$, $\sum_{j=1}^{|X_i|} \delta_{ij} = 0$ holds.

We note that each feature in $v$'s should be correlated with some features from $s_{ij}^{MSF}$ as otherwise such features can be dropped from $v$ without effecting selections.

The EM algorithm iteratively solves discrete choice models with temporary labeled requests and assigns "soft-labels" to unlabeled requests until convergence. In each iteration, temporary labeled requests are changed. The CL algorithm finds a partition of $X^L$ with a given number of clusters, then assigns labels to unlabeled requests using discrete choice models trained within each cluster, and eventually solves one model with all labeled requests. The XCL1 and XCL2 algorithms which have not yet been proposed for SSL start clustering with no fixed number of clusters, and develop partitions automatically. After the partitions have been found, the remaining steps are the same as in the CL algorithm.

2.2 EM algorithm

Let $\Theta$ be a coefficient vector. Then utility reads $U_{ij} = s_{ij}\Theta$ and the probability of choosing choice $j$ for request $i$ is:

$$prob(\delta_{ij} = 1) = \frac{e^{s_{ij}\Theta}}{\sum_{k=1}^{|X_i|} e^{s_{ik}\Theta}}.$$

Let $H = \{H_1, H_2, \ldots, H_Q\}$ be the set of all possible $\{L^*_{n+1}, L^*_{n+2}, \ldots, L^*_{n+m}\}$ for unlabeled requests, with $Q = \prod_{i=n+1}^{n+m} |X_i|$. Also let $q(\cdot)$ be the distribution function for hidden labels and denote $t$ as the iteration number and let $f(q, \Theta) = \sum_{l=1}^{Q} q(H_l)\log\left(\frac{P(D, H_l|\Theta)}{q(H_l)}\right)$.

In the E-step we compute $q(H_l)$ based on current $\Theta^t$ and find the "soft labels" for unlabeled data. This requires simple calculation based on utilities $U_{ij}$. Then we find $\Theta^{t+1}$ which maximizes $f(q, \Theta)$. This is the M-step and it gradually increases the lower bound of $\log(P(D|\Theta))$ until convergence. This step is the classic discrete choice utility coefficients computation algorithm based on maximum likelihood.

2.3 CL algorithm

In the CL, XCL1 and XCL algorithms, denote $K = \{1, 2, \ldots, |K|\}$ as the index set of the clusters. Let $C = \{C_1, C_2, \ldots, C_{|K|}\}$ be a set of clusters where $C_k$ is a subset of $I$ for any $k$ in $K$. Let $c^k$ be the centroid vector for cluster $C_k$. Given an arbitrary subset $C_k \subseteq V$, we denote by $D_k$ the matching subset corresponding to $C_k$ in $D^L \cup D^U$. Let $\ell(C_k)$ be the number of labeled requests in cluster $C_k$.

When applying the CL algorithm in choice models, we consider the individual specific features as the clustering features. An example is presented in Figure 1 to illustrate the clustering process. Suppose we have three requests, $X_1, X_2$ and $X_3$. From them, we draw the individual specific feature vectors as $V = \{v_1, v_2, v_3\}$ and partition on $V$. In this example, requests $X_1$ and $X_2$ have been clustered together and $X_3$ is a standalone cluster. Then based on $L_1, L_2$ and $L_3$, we generate the clustered data set $D_1$ and $D_2$ for choice modeling (considering in each cluster only requests from $D^L$).

$$X_1 \begin{Bmatrix} v_1 & s_{11}^{MSF} \\ \vdots & \vdots \\ v_1 & s_{1,|X_1|}^{MSF} \end{Bmatrix} \\ X_2 \begin{Bmatrix} v_2 & s_{21}^{MSF} \\ \vdots & \vdots \\ v_2 & s_{2,|X_2|}^{MSF} \end{Bmatrix} \Rightarrow V = \{v_1, v_2, v_3\} \Rightarrow C_1 = \{v_1, v_2\}, C_2 = \{v_3\} \Rightarrow \begin{matrix} D_1 \begin{Bmatrix} v_1 & s_{11}^{MSF} & \delta_{11} \\ \vdots & \vdots & \vdots \\ v_1 & s_{1,|X_1|}^{MSF} & \delta_{1,|X_1|} \\ v_2 & s_{21}^{MSF} & \delta_{21} \\ \vdots & \vdots & \vdots \\ v_2 & s_{2,|X_2|}^{MSF} & \delta_{2,|X_2|} \end{Bmatrix} \\ D_2 \begin{Bmatrix} v_3 & s_{11}^{MSF} & \delta_{31} \\ \vdots & \vdots & \vdots \\ v_3 & s_{3,|X_3|}^{MSF} & \delta_{3,|X_3|} \end{Bmatrix} \end{matrix} \\ X_3 \begin{Bmatrix} v_3 & s_{31}^{MSF} \\ \vdots & \vdots \\ v_3 & s_{3,|X_3|}^{MSF} \end{Bmatrix}$$

Figure 1. An illustrative example for choice model clustering

In the CL algorithm, we start with a fixed number of clusters $K$ and partition $V$ into $K$ clusters. Clustering of $C$ implies clustering of $D$. Let us denote $D = \{D_1, D_2, ..., D_K\}$ as the resulting partition of $D$. Then $D_k^L = D_k \cap D^L$ and $D_k^U = D_k \cap D^U$. We calibrate a choice model for each $D_k^L$ and get the coefficient vector $\Theta_k$ for cluster $k$ which also imply labels in $D_k^U$. The entire CL algorithm enumerates all possible $K$ and the best performed one is chosen. To guarantee the computation of "soft labels," it is required that the number of labeled choice set in the $kth$ cluster is greater than the minimum number of requests to calibrate the model. For details of the CL algorithm, we refer to Zhu and Goldberg (2009).

2.4 XCL1 algorithm

While EM and CL are standard algorithms in SSL that we have adapted to discrete choice, the remaining two algorithms are derivations of CL that to the best of our knowledge have not yet been observed in the past (as such they are applicable in the context of general SSL.) Distinct from the CL algorithm, the XCL1 algorithm requires no number of clusters as an input. It develops the clusters automatically without a target number of clusters. A diagram representing the clustering process is displayed in Figure 2. In step (a), we partition $V$ into two clusters $C_1$ and $C_2$ and obtain two centroids (the crosses in Figure 2). In step (b), the algorithm develops two random vectors passing through $C_1$ and $C_2$'s centroids with the length of each vector being equal to the average within cluster distance. We use the end points as the new centroids to further partition clusters $C_1$ and $C_2$ into two. In step (c) with BIC defined appropriately, it is assumed that since $bic(D_1, D_2, \Theta_1, \Theta_2)$ in step (c) is greater than $bic(D_1, \Theta_1)$ in step (b), and $bic(D_3, D_4, \Theta_3, \Theta_4)$ in step (c) is greater than $bic(D_2, \Theta_2)$ in step (b), we have four new clusters developed from the original $C_1$ and $C_2$. Otherwise, if the new BIC is less than the old one, then the old cluster would not be partitioned into two. The algorithm then repeats steps (b) and (c) to partition each of the new clusters until no more partitions can be developed. If at any point the number of labeled requests with a cluster is less than parameter $m$, we discard such a cluster.

BIC of $K$ clusters in the semi-supervised problem is defined as:

$$bic(D_1,\ldots,D_K,\Theta_1,\ldots,\Theta_k) = \sum_{k=1}^{K}\log\left(P\left(D_k^{U^*},D_k^L|\Theta_k\right)\right) - \frac{K\cdot R}{2}\log\left(\sum_{k=1}^{K}|D_k|\right).$$

Here $R$ is the number of coefficients in the choice model (each cluster has the same coefficients) and $|D_K|$ the number of requests in cluster $k$. We refer to Algorithm 1 for details.

In the very first iteration, the computation of BIC starts with a model estimation based on $D_k^L$ and then "soft labels" are assigned to $D_k^U$ according to the coefficients of the model. After that, we combine $D_k^L$ and $D_k^{U^*}$ together to re-estimate the model and get the BIC.

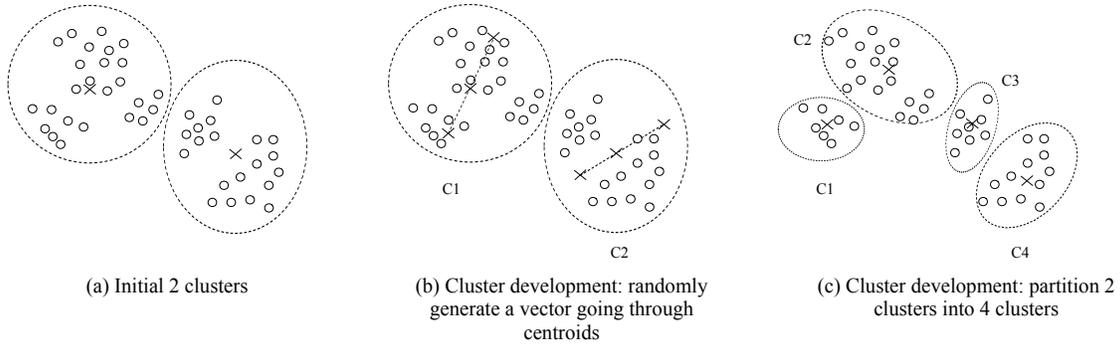

(a) Initial 2 clusters    (b) Cluster development: randomly generate a vector going through centroids    (c) Cluster development: partition 2 clusters into 4 clusters

Figure 2. Illustration of XCL1 algorithm

---

Algorithm 1. XCL1

Input: $D^L, D^U$; output: $\Theta$

Initialize: $K=2$; maximum number of clusters $K_{max}$; partition $\{X^L \cup X^U\}$ into $R_C = \{C_1, C_2\}$ based on $V$; minimum number of iterations for cluster split $IterNumMax$

1: While $K < K_{max}$ do
2:    R = ∅ // new clusters
3:    For each $C_k$ in $R_C$
4:       Let $IterNum = 0$
5:       While $IterNum < IterNumMax$ do
6:          Based on uniform distribution between [-1,1], generate a random vector $g$
7:          Compute $\hat{g} = \frac{g}{||g||}$
8:          $\overline{u_k} = \frac{1}{|C_k|}\sum_{v_i \in C_k}||v_i - c^k||$
9:          $c_{new} = \{c^k + \overline{u_k}\hat{g}, c^k - u_k\hat{g}\}$
10:         Partition $C_k$ into $C_{k1}$ and $C_{k2}$ by using 2-means with $c_{new}$ as starting centroids
11:         If $\ell(C_{k1}) > m$ and $\ell(C_{k2}) > m$
12:            Break
13:         End
14:         $IterNum += 1$
15:       End While
16:       If $\ell(C_{k1}) > m$ and $\ell(C_{k2}) > m$
17:          Compute coefficients $\Theta_{k1}, \Theta_{k2}$ for $D_{k1}, D_{k2}$
18:          If $bic(D_{k1}, D_{k2}, \Theta_{k1}, \Theta_{k2}) < bic(D_k, \Theta_k)$
19:            $R = R \cup \{C_k, (C_{k1}, C_{k2})\}$
20:    End For
21:    If $R = \emptyset$
22:       Break
23:    Else
24:       Update $R_c$ by using $R$
25:    End
26: End While
27: Compute $\Theta$ based on $R_C$

---

## 2.5 XCL2 algorithm

Instead of developing each cluster independently, XCL2 aims at partitioning a pair of clusters into three clusters starting from one more potential centroid. In Figure 3, step (a) is the starting point with 2 clusters which can be obtained as in Algorithm 1 in an iteration. Once Algorithm 1 terminates as it can no longer split clusters, we start a new partitioning direction. In the example below, suppose Algorithm 1 does not split the two clusters. Then we invoke steps (b) and (c) in Figure 3. Step (b) first randomly selects two clusters and constructs a connecting vector (the dotted line) between the two clusters' centroids (e.g. centroid 1 and centroid 2). Then it computes a new centroid (e.g. centroid 3) which lies in the middle of the connecting vector. In step (c), the algorithm partitions the data by 3-means into three clusters based on the three initial centroids (centroid 1, centroid 2 and centroid 3) if and only if $bic(D_1, D_2, \Theta_1, \Theta_2)$ in step (b) is less than $bic(D_1, D_2, D_3, \Theta_1, \Theta_2, \Theta_3)$ in step (c). If the algorithm can successfully split the initial two clusters into three, it then invokes Algorithm 1, followed by steps (b) and (c) to further split into three clusters until no more partitions can be developed. Details of the XCL2 algorithm are presented in Algorithm 2.

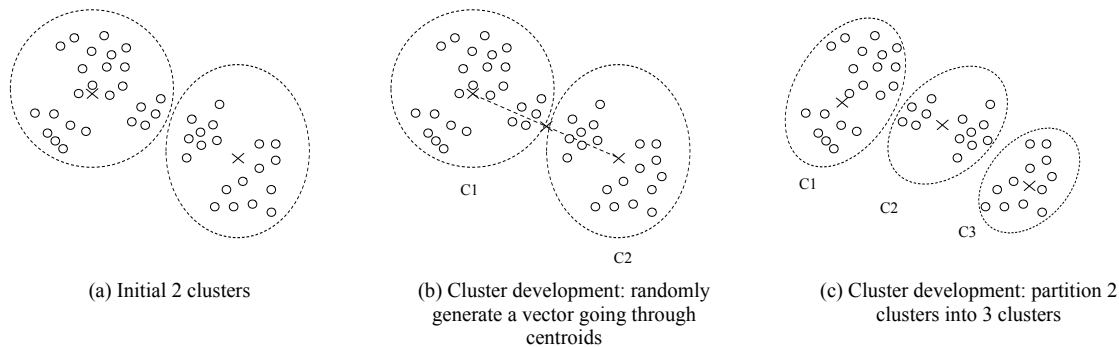

(a) Initial 2 clusters  
(b) Cluster development: randomly generate a vector going through centroids  
(c) Cluster development: partition 2 clusters into 3 clusters

Figure 3. Illustration of XCL2 algorithm

---

Algorithm 2. XCL2

Input: $D^L, D^U$; output: $\Theta$

Initialize: $K=2$; maximum number of clusters $K_{max}$; partition $\{X^L \cup X^U\}$ into $R_C = \{C_1, C_2\}$ based on $V$; denote by $CP$ the set containing all pairs of clusters from $R_C$; minimum number of iterations for cluster split $IterNumMax$

1: While $K < K_{max}$ do  
2:    Execute Algorithm 1  
3:    If $R_C$ has not been updated  
4:       $R = \emptyset$  
5:       For each $\{C_{k1}, C_{k2}\}$ in $CP$  
6:          $c^{k3} = \frac{1}{2}(c^{k2} + c^{k1})$  
7:          Partition $C_{k1} \cup C_{k2}$ by using 3-means into $\{C_p, C_q, C_o\}$ with $\{c^{k1}, c^{k2}, c^{k3}\}$ as starting centroids  
8:          Compute coefficients $\Theta_p, \Theta_q, \Theta_o$ for $D_p, D_q, D_o$  
9:          If $\ell(C_p) > m, \ell(C_q) > m, \ell(C_o) > m, bic(D_p, D_q, D_o, \Theta_p, \Theta_q, \Theta_o) < bic(D_{k1}, D_{k2}, \Theta_{k1}, \Theta_{k2})$  
10:            $R = R \cup \{(C_{k1}, C_{k2}), (C_p, C_q, C_o)\}$  
11:            Break // we break once a better partition is found to save computational time  
12:       End For  
13:       If R = $\emptyset$  
14:          Break  
15:       Else  
16:          Update $R_c$ by using $R$  
17:       End  
18: End While  
19: Compute $\Theta$ based on $R_C$

---

## 3. Computational Studies

In order to evaluate the performance of the algorithms, we used a labeled hotel data set and removed the labels from a subset of the data. We applied the ROL model in this case. The reason we apply the ROL model is that it proliferates the number of choice sets in our hotel data set, and as a benchmark study the hotel data set is relatively small compared to the airline data set presented later. As the ROL model can be transformed into a series of MNL models, the formulations and algorithms in Section 2 hold. In the second case study, we analyzed an airline data set by manually labeling a small subset with a rule. The proposed algorithms were then applied in this case. Prediction accuracies are provided.

3.1 Hotel case study

We start with the hotel study where all of the choices are available. A data description and experimental design, performance analyses and algorithm recommendation are presented below.

3.1.1 Data description and experimental design

Data was collected from five U.S. properties of a major hotel chain for check-in dates between March 12th, 2007, and April 15th, 2007 (Bodea et al., 2009) and made publicly available. We refer to Bodea et al. (2009) for details about the data format and descriptive statistics. In total, there are 3,140 requests. Since each subsidiary of this hotel chain has different names for a similar service, we aligned the service names based on the descriptions of each hotel's service packages. For example, "accommodation combined with in city activities" and "accommodation combined with special event activities" are aligned with the name "accommodation with activities." On the other hand, some hotels have specific descriptions of room type, such as "non-smoking queen beds room" or "smoking queen beds room." We simply grouped them based on their upper level description which is their room type (e.g. queen beds room). For details about feature descriptions, see Tables 1 and 2 in the appendix.

The data is set to calibrate classic choice models such as the MNL models instead of the ranked choice models. So we first transformed this data set into a ranked data set. To convert, we estimate the MNL models' coefficients first and use them to re-estimate each choice utility and also the rank. With the ranked data, we applied the ROL model instead of the traditional MNL model. To evaluate the performance of the predicted rank, four metrics were applied: log-likelihood for ranked data, rank difference (e.g. predicted rank against the true rank), position difference and reciprocal rank. The descriptions are summarized below.

*Rank-ordered log-likelihood (ROLIK)*. In *ROLIK*, we roll up the log-likelihood of all choices within each choice set while in MNL we only consider the chosen choice. Denote $r_i$ as the rank of choices in $X_i$. According to Fok et al. (2007), the log-likelihood of rank $r_i$ in choice set $X_i$ is:

$$\log(P(r_i|\Theta)) = \log\left(P(U_{i1} > U_{i2} > \cdots U_{i|X_i|})\right) = \log\left(\prod_{j=1}^{|X_i|-1} \frac{exp(U_{ij})}{\sum_{j=i}^{I} exp(U_{ij})}\right).$$

A better model yields a higher *ROLIK* value.

*Rank difference (RD).* Kendall's $\tau$ is a non-parametric estimator of the correlation between two vectors of random variables. Let $\tau(x, y)$ be the Kendall's $\tau$ between rank $x$ and $y$. Denote $r^T, r^E$ and $r^R$ as the true rank, the estimated rank and a randomly generated rank, respectively. Since Kendall's $\tau$ is between $[-1, 1]$, we capture $\frac{1-\tau(r^T, r^E)}{2}$ and $\frac{1-\tau(r^T, r^R)}{2}$ which are within $[0, 1]$. For this metric, a lower value indicates that the two ranks are similar.

*Position difference (PD).* Given choice set $X_i$, denote $j_{(i)}$ as the index in the original data set with $\delta_{ij_{(i)}} = 1$. Let $r_i^E$ be the estimated rank and $p$ be the estimated rank of choice $j_{(i)}$ in $r_i^E$. We define $PD$ as $\frac{p-1}{|X_i|-1}$. Using this metric, a better model yields a lower value.

*Reciprocal rank (RR).* The *mean reciprocal rank (MRR)* for a set of $Q$ queries is defined as $MRR = \frac{1}{|Q|} \sum_{i=1}^{|Q|} \frac{1}{p_i}$ where $p_i$ is the rank of the first correct response in every $i$ in returned responses. In our case, $p_i$ is the rank of the chosen choice in the original data set. Using this metric, a better model generates a higher value.

The goal of the experiment is to evaluate each algorithm's predicting accuracy under different labeled percentages. So the experimental design includes two parts: (1) transforming the original data set into a ranked data set; (2) designing the 10-fold cross validation experiment and the baseline. We use Label-$q\%$ to indicate the experiment with $q\%$ labeled data and $(100 - q)\%$ unlabeled data. We transformed the original data set since this allowed us to apply the ROL model. We used cross validation since it provided a balanced evaluation of algorithms' prediction and reduced variability. With high $q\%$, the marginal benefits of applying SSL may be reduced so we focus on Label-10% to Label-60% cases in increments of 10%. In each cross validation step we want the ranks of labeled requests to be consistent, for example, Label-20% uses labeled data from Label-10% which should have consistent ranks in requests appearing in both. In order to achieve this property, we proceed as follows. We have first evenly divided the data set into 10 batches (see Figure 4) so each batch represents 10% of data. We have then randomly selected 4 batches and removed their labels as they would be unlabeled throughout the experiment. The next step is to train an MNL model on each of the remaining 6 batches independently. We use each batch's coefficients to compute each choice's utility and rank them within each request. If we want to conduct the experiment Label-20%, we draw the ranked batches (i.e. the first and second batches) and remove the rank labels in the remaining 4 batches (see Figure 5).

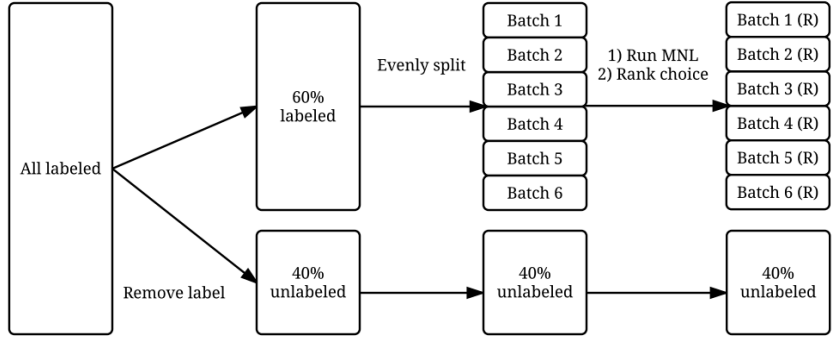

Figure 4. Experimental design: generation of ranked data

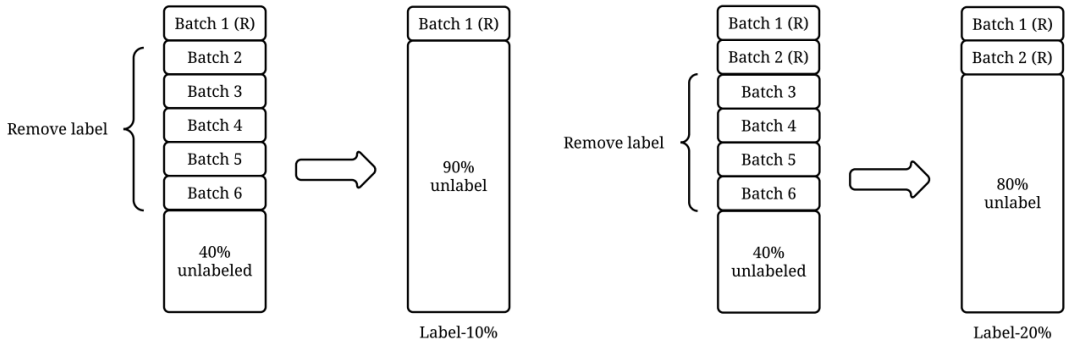

Figure 5. Experimental design: Label-10% and Label-20% cases

To measure the prediction accuracy, we first explain the use of train and test sets. Since we want the test set to be consistent across all experiments, we always select it from batch 1 (which is present in all experiments). In a single fold experiment, we split each batch into 10% and 90%. All 90% sets are used for training (in addition to all unlabeled data) and the 10% from batch 1 is used as test. This is repeated 10 times over the 10 different folds.

The baseline model for an experiment consists of the ranked logit model calibrated on all labeled train instances.

The model estimation part has been coded in Spark Scala using the stochastic gradient descent (SGD) method. The cluster includes four 2.2GHz Xeon CPUs with each of them having 8 cores and 32 GB memory. Cloudera version CDH 5.8 is the underlying Hadoop distribution. After parameter tuning for all combinations of step size (from 1 to 40 with increment of 1) and sampling rate (from 0.2 to 0.8 with increment of 0.2), we found step size=40 and sampling rate=0.2 yielded the highest log-likelihood. Spark uses the learning rate of $\frac{Step\ size}{\sqrt{Number\ of\ iterations}}$. So we used this setting for the hotel case. The EM algorithm stops when the coefficient change $\left\|\frac{1}{t+1}\sum_1^{t+1}\Theta^{t+1} - \frac{1}{t}\sum_1^t \Theta^t\right\| < 0.01$. In the EM algorithm, due to the computational cost of the 10-fold cross validation steps and model estimation (the EM algorithm converges slowly), we randomly draw $\beta$ percent requests from $D^U$, and assign them the "soft-labels" in each iteration $t$ and ran ranked logit model on the current labeled data and random draw. To implement the CL and XCL algorithms, we use the Lloyd algorithm in the clustering process as it can assign starting centroids. K-means and Euclidean distance are used in the clustering process.

3.1.2 Performance analyses

In this performance analyses section, Figures 6 and 7, we compare the four metrics computed from the proposed algorithms to the baseline prediction. All metric values are compared to the baseline and numbers are presented in percentage. For example, in the Figure 6's upper chart, the y-axis represents $\frac{ROLIK-Baseline}{Baseline} \times 100\%$. For the CL algorithm, we experimented with the different number of clusters (from 2 to 6 with increments of 1) and report only the best one. Similarly, in the EM algorithm, we experimented with two $\beta$ percent requests (i.e. 5% and 10%) and report only the best one in the figure. With regard to Kendall's $\tau$ against a random rank, we observed (results not shown here) that the values for all experiments were around 0.5 which means $\tau(r^T, r^R)$ tended to be zero. It indicates that the results are distant from random decisions. The main discoveries are summarized in Table 3.

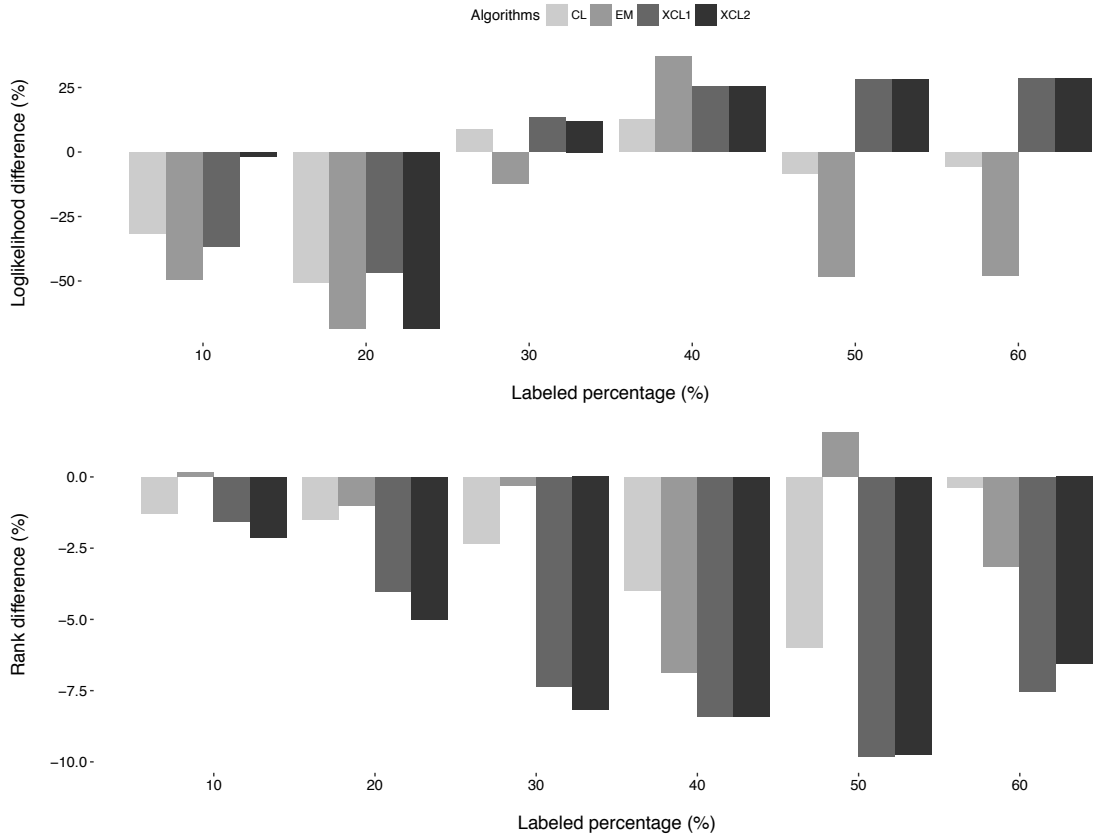

Figure 6. Prediction accuracy plots for log-likelihood and rank difference

Table 3. Experiment analysis-1

| Metric | Phenomena | Insights |
|---|---|---|
| ROLIK | When labeled percentage is low ($q\% < 30\%$), all algorithms yield negative values. | (1) When using ROLIK as the criterion, the SSL-DCM algorithms perform better in low labeled percentage scenarios.<br>(2) The algorithms work best at 20%. |
| ROLIK | When labeled percentage is high ($q\% \geq 30\%$), algorithms yield more positive values. | Using ROLIK, it indicates that the algorithms are not stable in finding relative utility differences between choices when labeled percentage is high. In one case when SSL and baseline both predict rank correctly, SSL's ROLIK may not be higher than the baseline since the probability depends on utility differences. In another case, even though SSL predicts a better rank than the baseline, it is still not guaranteed its ROLIK is higher. A simple example is presented in Table 4 showing that even when the semi-supervised prediction can predict the correct ranking ($U_3 > U_2 > U_1$) and the baseline predicts incorrectly ($U_2 > U_3 > U_1$), its ROLIK measurement may not be higher than the baseline. |
| ROLIK | When labeled percentage is high ($q\% \geq 30\%$), the EM algorithm generates negative results. | Using ROLIK, the EM algorithm is more stable with respect to the labeled percentage. This is reasonable since Bayesian approaches rely on distribution fitting while clustering based algorithms are more about distance minimization. |
| ROLIK | Except for $q\% = 40\%$, the EM algorithm generates the lowest values. | Using ROLIK, the EM algorithm outperforms other algorithms in most situations. |
| RD | 10% to 50%:<br>$q\% \uparrow$ implies $RD \downarrow$ and on average $RD < 0$<br>50% to 60%:<br>$q\% \uparrow$ implies $RD \uparrow$ and on average $RD < 0$ | (1) When using RD as the predicting criterion, the algorithms have excellent performance as almost all results are better than the baseline.<br>(2) In general, the algorithms have a much better prediction power when $q\%$ is below a certain level (e.g. 50%). |
| RD | Clustering based algorithms (i.e. CL, XCL1 and XCL2) are always below zero while the EM algorithm is above zero at 50%. | When using RD as the predicting criterion, the clustering based algorithms work better and are more stable than the EM algorithm. |
| RD | XCL2 is lower than XCL1 before 40% but higher than XCL1 after 40%. | When using RD as the predicting criterion, XCL2 is better in low labeled scenario. But when labeled data grows, the power diminishes. |

Table 4. An example displaying *ROLIK* prediction

| True rank | Baseline utility | SSL utility |
|---|---|---|
| Choice 3 | $U_3 = 4$ | $U_3 = 5$ |
| Choice 2 | $U_2 = 5$ | $U_2 = 3$ |
| Choice 1 | $U_1 = 1$ | $U_1 = 2$ |
| *ROLIK* | $\log\left(\frac{4}{10} * \frac{5}{6}\right) \approx \log(0.33)$ | $\log\left(\frac{5}{10} * \frac{3}{5}\right) = \log(0.30)$ |

For the metric *PD* and *RR* (see Figure 7), the trend varies. The pattern in different SSL-DCM algorithms is not consistent with what is observed in log-likelihood and rank difference. For example, the algorithms generally perform better in Label-40% than in Label-50% but are worse than baseline in other labeled percentage for metric *PD*. A possible reason accounting for the inconsistent pattern is that this metric measures the position of the originally chosen choice in the estimated rank, however the ROL model is to maximize the log-likelihood of all ranked choices. We also categorize some phenomena and insights in Table 5.

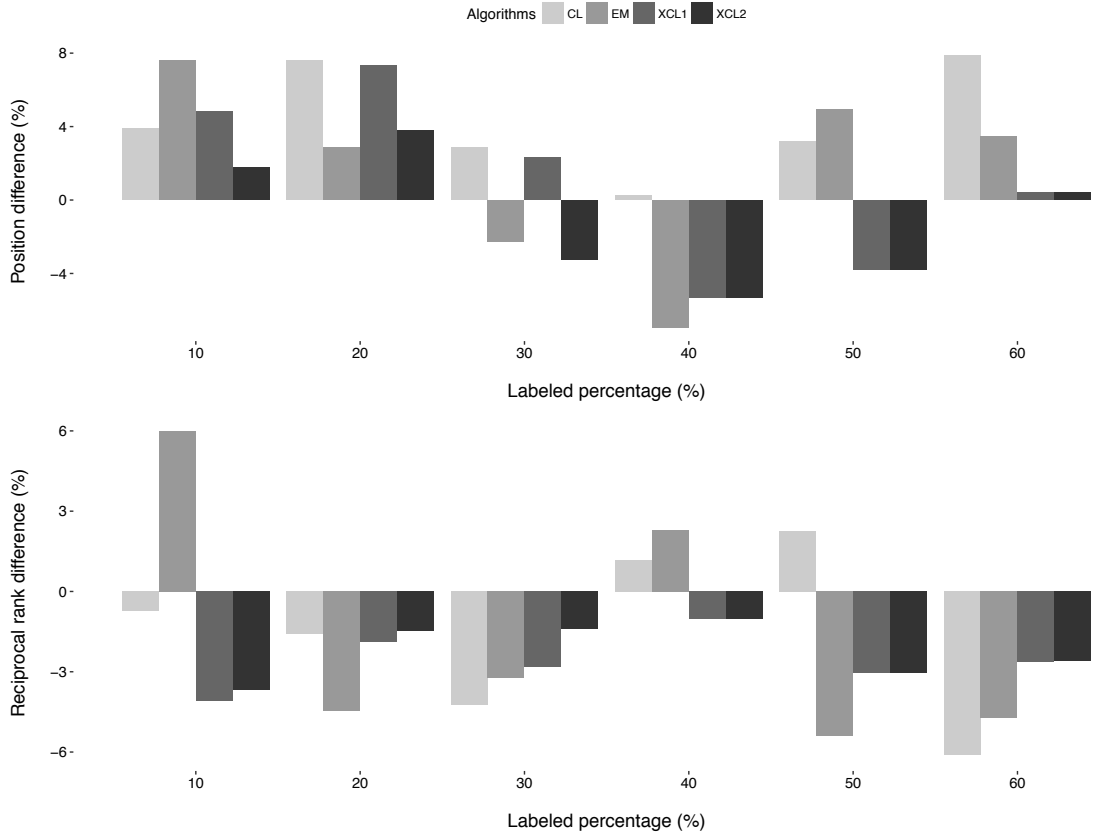

Figure 7. Prediction accuracy plots for position difference and reciprocal rank

Table 5. Experiment analysis-2

| Metric | Phenomena | Insights |
|--------|-----------|----------|
| *PD* | XCL2 is lower than XCL1 before 40% but similar to XCL1 after 40%. | Similar to *RD*, the XCL2 algorithm works better than XCL1 only when $q\%$ is less than a certain value. |
| *PD* | XCL2 always generates the lowest or second lowest values. | Using *PD*, the XCL2 algorithm works better than other algorithms. |
| *RR* | XCL2 is higher than XCL1 before 40% but similar to XCL1 after 40%. | Same as above. |

3.1.3 Algorithm recommendation

To provide recommendations for using the algorithms, we first analyze their computational times which are displayed in Figure 8. The figure shows the combined running time for each algorithm (e.g. for the CL algorithm, we add the running time of all possible clusters together). In Figure 8, when labeled percentage is below 20%, the EM algorithm is not the most time consuming method, yet when $q\% \geq 20\%$ its running time grows quickly. This indicates the EM algorithm may not fit large $q\%$. Meanwhile, CL spends similar time compared to XCL1. When $q\%$ is small (e.g. 10% or 20%), the time difference between XCL1 and XCL2 is larger than a lager $q\%$ (e.g. 30%-60%).

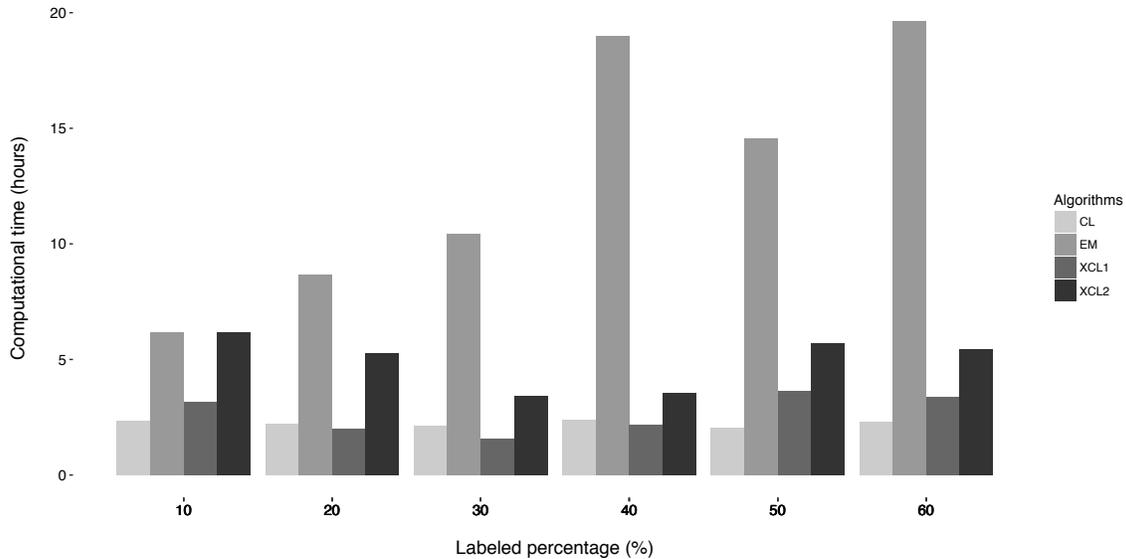

Figure 8. Combined computation times for SSL-DCM algorithms

Based on the previous prediction accuracy and computational time displayed in Figure 8, in Figure 9 we provide recommendations for using the algorithms under different scenarios. The x-axis represents the computation time while the y-axis represents the difference between algorithms' prediction and the baseline by using the rank difference metric. We use this metric since it considers the relations among choices in a choice set. For x-axis and y-axis, a lower value indicates a better algorithm. If the data set contains 10% labeled data, we recommend to use CL if there is time sensitivity and XCL1 if higher prediction accuracy is required. If the data set contains 30% or 50% labeled data, we recommend to use XCL1 if there is time sensitivity and XCL2 if high prediction accuracy is required. Based on this, we discover that the XCL1 algorithm's prediction accuracy is better and more consistent when labeled data increases. Label percentages 20%, 40%, 60% are not displayed in Figure 9 but we observe that XCL1 always lies in the efficient boundary (e.g. the dotted lines in Figure 9), and for 40% and 60% it yields best prediction accuracy.

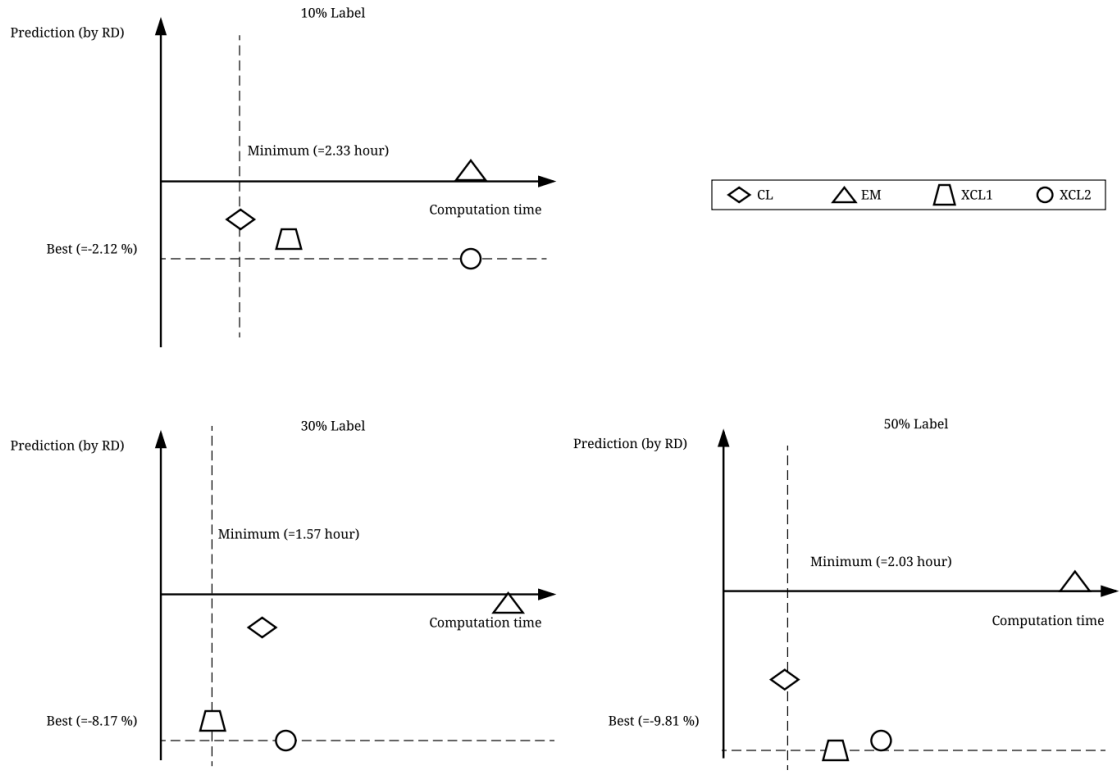

Figure 9. Algorithm recommendations for different scenarios

3.2 Airline shopping case study
3.2.1 Data introduction

All of the demand and itinerary data were from a GDS provider. The case is based on a single international market with travelers' departure dates spanning one year. Since the entire data set is large, we only focused on round-trips. We use the term "service" for one or multiple flight legs connecting travelers' origin and destination. So one round-trip includes two services. Since we were not provided the actual bookings made by travelers, we came up with a heuristic method to infer some of the bookings. We have 11,000 requests and almost 2 million itineraries returned by the GDS. According to a survey conducted by Ipsos Public Affairs (2016), "Total travel price" was ranked as important by 86% of those respondents who flew in 2015 and the next three most cited factors were airline schedule (83%) and total travel time (77%). Inspired by this fact, we made a modest assumption that a traveler purchases the itinerary with the lowest fare and the best match to his or her travel preferences. To measure the best match of travelers' preference, we first define quality $H_{ij}$ of itinerary $j$ for request $i$.

$$H_{ij} = \left|t_i^{d1} - t_j^{d1}\right| + \left|t_i^{d2} - t_j^{d2}\right| + \left|t_i^{a1} - t_j^{a1}\right| + \left|t_i^{a2} - t_j^{a2}\right| + \left|e_i - e_j\right|$$

$t_i^{d1}, t_j^{d1}$ – Requested departure time of the first service and itinerary's departure time of the first service

$t_i^{d2}, t_j^{d2}$ – Requested departure time of the second service and itinerary's departure time of the second service

$t_i^{a1}, t_j^{a1}$ – Requested arrival time of the first service and itinerary's arrival time of the first service

$t_i^{a2}, t_j^{a2}$ – Requested arrival time of the second service and itinerary arrival time of the second service

$e_i, e_j$ – Elapsed time of request $i$ and itinerary $j$, respectively.

A lower $H_{ij}$ yields a better quality of response. Let $F_{ij}$ be the fare of itinerary $j$ in response to request $i$ and let $LF_i$ be the lowest fare in request $i$. Similar to $LF_i$, $LH_i$ is defined as the lowest $H_{ij}$. For $\delta_{ij}$, we set $\delta_{ij} = 1$ if $H_{ij} = LH_i$ and $F_{ij} = LF_i$ and $\delta_{ij} = 0$, otherwise.

The assumption is that if there is an itinerary with lowest fare and lowest deviation from preference, then it is booked. If $\sum_{j=1}^{|X_i|} \delta_{ij} = 1$, then we assume request $i$ is labeled. If not, the request is unlabeled. In our data set we have 1,000 labeled requests and 10,000 unlabeled requests.

3.2.2 Computational analyses

With the airline data, we know travelers' underline preferences for airlines. But in the experiment, we left them out from the quality formula since it is a categorical feature and unrelated to time. In the logit model, we consider features shown in Appendix Table 7 related to the calculation of $H_{ij}$ and $F_{ij}$. In order to verify the prediction accuracy, we expect the yielded model coefficients would select an itinerary with low fare and good quality of response. Besides the intercept, the model considers three alternative specific features. The first feature $f_{ij}^1$ captures the gap between each itinerary $j's$ fare and the average fare in $X_i$, formally $f_{ij}^1 = F_{ij} - \frac{1}{|X_i|}\sum_{j=1}^{|X_i|} F_{ij}$. We apply this feature instead of using the fare directly because the fare range of different requests may vary a lot, even in the same market, and different advanced purchase days may contribute to distinct fare levels. The second and third features are defined as $f_{ij}^2 = t_i^{d1} - t_j^{d1}$ and $f_{ij}^3 = t_i^{d2} - t_j^{d2}$, respectively. Ideally we should include the calculation of requested elapsed time and requested arrival time in the calculation of $H_{ij}$, but both fields are empty in most of the records, and they were left out. In addition, we did not compute the absolute value for $f_{ij}^2$ and $f_{ij}^3$ because compared to a traveler's desired time, an earlier departure is different from a later departure. On the other hand, in the formula of $H_{ij}$ we consider the absolute values of the differences since we do not want the negative values offset the positive values.

Similar to the hotel case, we used the SGD method in Scala since the logit model estimation was computationally expensive and it took more than 120 hours in R to estimate the model with the entire data set. The input data was stored in HDFS and the cluster environment was the same as in the hotel case. We also tuned the SGD parameters by trying combinations of step sizes and sampling rates and found step size=7 and sampling rate=0.8 yielded the highest log-likelihood. So we use this setting for the airline case.

Due to confidentiality, we do not present the exact coefficients of SSL-DCM here. Since we have coefficients for fare and also time related features, we can estimate the value of time (i.e. dollars per hour). The coefficients of fare gap and departure time have units $utility/dollar$ and $utility/hour$, respectively. So we simply compute the ratio of these two and get the value of time. We discover that travelers value more the first departure time than the second departure time. For example, the EM algorithm indicates that the value of the first departure time is between $250-$300 per hour while the value of the second departure time is between $150-$200 per hour.

We designed two metrics to evaluate the prediction accuracies of the coefficients. As mentioned before due to the strategy of labeling the 1,000 requests, we expect the coefficients to predict an itinerary with the lowest fare and the best match. Let us first define the term "lowest $p\%$ records." In each $X_i$, we sort the itineraries by $F_{ij}$ in ascending order and find $p\%$ itineraries with lowest fare, denoted as $A_i$. In a similar way, we sort the itineraries by $H_{ij}$ for each $X_i$ and find $p\%$ itineraries with lowest $H_{ij}$, denoted as $B_i$. Let $Z_i$ be $A_i \cup B_i$. Suppose $F_i^p$ is the fare and $H_i^p$ is the quality of the predicted itinerary of request $i$. Then for each request $i$, we define the prediction accuracies as $MF_i = (F_i^p - \frac{1}{|Z_i|}\sum_{j=1}^{|Z_i|} F_{ij})/\frac{1}{|X_i|}\sum_{j=1}^{|X_i|} F_{ij}$ and $MH_i = (H_i^p - \frac{1}{|Z_i|}\sum_{j=1}^{|Z_i|} H_{ij})/\frac{1}{|X_i|}\sum_{j=1}^{|X_i|} H_{ij}$. For these metrics, a lower value indicates a better prediction.

We tested the metrics with $p\%$ ranging from 10% to 50% with the step size of 10%. The results are presented in Table 6. From Table 6, we observe that the generated coefficients have a good prediction in terms of fare. For example, both the CL and XCL1 or XCL2 algorithms are better than the average when $p\% \geq 10\%$. However, with respect to response quality, only the clustering based algorithms are slightly close to the benchmark and this requires $p\% = 50\%$. One explanation for this phenomena is that $f_{ij}^2$ and $f_{ij}^3$ in the $H_{ij}$ formula are captured by their absolute values which could lead to some discrepancies in terms of prediction accuracy.

Table 6. Prediction accuracies

|  |  | 10% | 20% | 30% | 40% | 50% |
|---|---|---|---|---|---|---|
| $MF_i$ (in %) | CL | 1.71 | **-1.68** | **-4.08** | **-5.98** | **-7.66** |
|  | EM | 29.70 | 24.81 | 21.79 | 19.28 | 17.13 |
|  | XCL1 | **-3.91** | **-7.10** | **-9.26** | **-11.08** | **-12.72** |
|  | XCL2 | 2.18 | **-1.49** | **-3.75** | **-5.69** | **-7.44** |
| $MH_i$ (in %) | CL | 35.47 | 26.25 | 15.35 | 8.49 | 3.02 |
|  | EM | 56.62 | 45.80 | 32.81 | 24.36 | 17.94 |
|  | XCL1 | 36.11 | 26.13 | 15.99 | 9.76 | 4.42 |
|  | XCL2 | 37.66 | 27.12 | 16.63 | 10.18 | 4.65 |

To better understand the XCL algorithms, we present the clustering development tree of the XCL1 algorithm (see Figure 10). The algorithm first partitions the original data into two clusters. Then each cluster is not able to be further partitioned. For one branch it is because the combined BIC of the potential split is lower than its parent node. For another branch, the algorithm cannot find enough number of observations in each cluster to guarantee the model estimation.

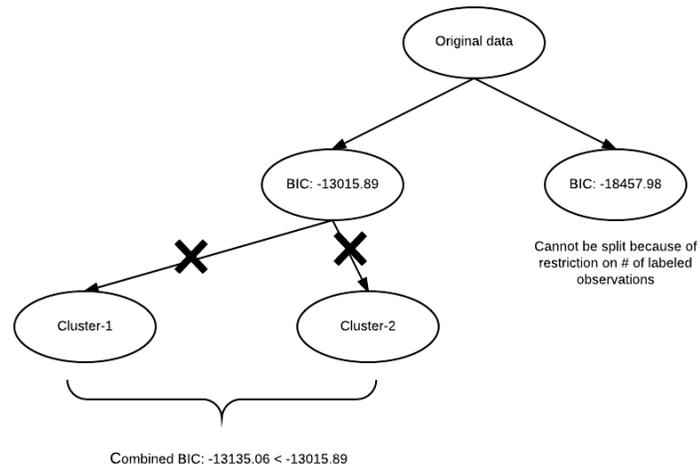

Figure 10. Clustering development in XCL1

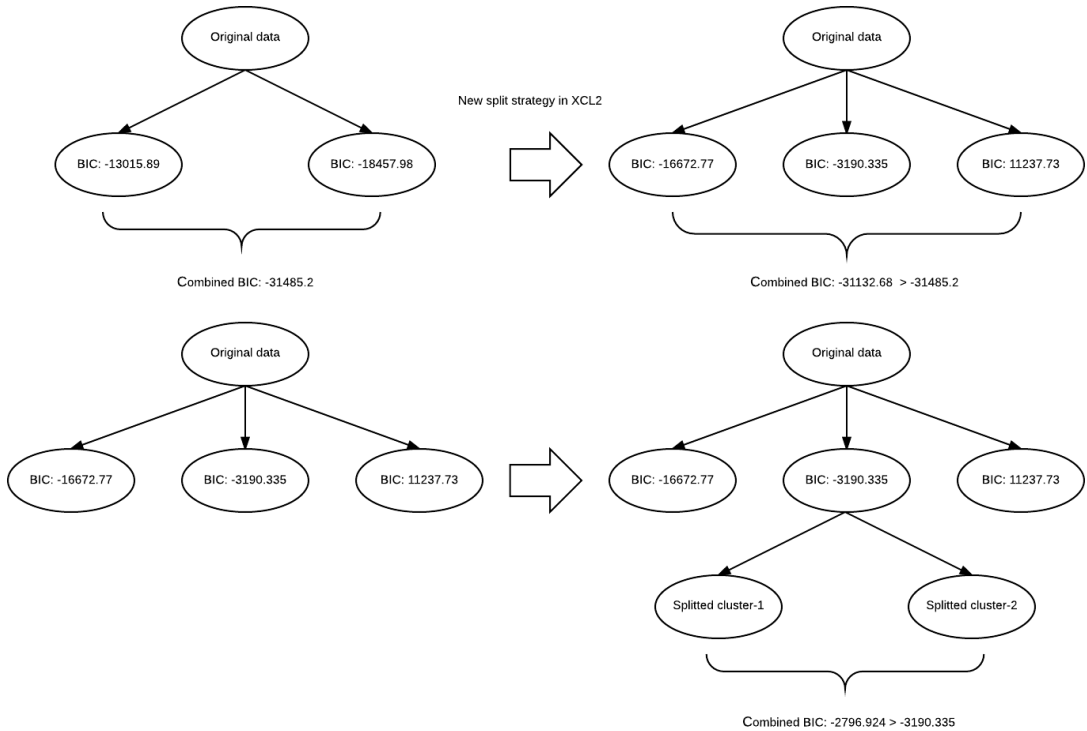

Figure 11. Clustering development in XCL2

Figure 11 shows how the XCL2 develops new clusters based on the initial two clusters from XCL1. First of all, XCL2 split the two clusters into three by initializing a new center lying between original centers as the combined BIC for three clusters in higher than the combined BIC for the original two clusters. Then it further split the middle cluster into two because the BIC criterion holds. Later, the algorithm stopped since no more splits could be found.

## 4. Conclusion

The SSL-DCM algorithms discussed in this paper explore new approaches to estimate discrete choice models. The algorithms can solve problems with limited labeled requests and many unlabeled requests. Besides adapting classic SSL algorithms such as EM and CL, we design two new algorithms XCL1 and XCL2 which can automatically segment the requests with an adapted BIC criterion. The SSL-DCM algorithms have been tested on two case studies. In the hotel case, we evaluated the prediction accuracies of the SSL-DCM algorithms using 10-fold cross validations. The main findings from this case study are summarized as follows.
- In general, the XCL1 algorithm has consistently good prediction accuracy and relatively low computation time.
- When the data set contains around 10% labeled data, we recommend to use CL if satisfactory solutions must be obtained quickly, and XCL1 if high prediction accuracy is needed.
- When the data set contains 20%, 30% or 50% labeled data, we recommend to use XCL1 if computational time is at a premium and prediction accuracy can be slightly sacrificed, and XCL2 if high prediction accuracy is needed.
- When the data set contains 40% or 60% labeled data, we recommend to use XCL1.

In the airline shopping case, we explored the travelers' purchasing behavior with a real data set from a GDS provider. The main findings are as follows.
- Results from the EM algorithm shows that travelers' value of time for first departure time is between $250-$300 per hour while the value of second departure time is between $150-$200 per hour.
- With regard to prediction accuracy, XCL1 outperforms CL, EM and XCL2 in this case.
- XCL2 is no better than XCL1 in terms of prediction accuracy even if finds more possible partitions in this case.

Future studies include improving the current SSL-DCM algorithms to yield a lower computation time and also to test them in more complex discrete choice models such as nested logit or cross-nested logit models.

## Acknowledgement

We are grateful to Sabre Holdings for providing the airline shopping data. Jie Yang has also been partially supported by a fellowship from the Transportation Center at Northwestern University.

# Appendix

Table 1. Description of hotel data set

| Name | Description |
| --- | --- |
| Booking ID | ID of a booking |
| Product ID | ID of a product |
| Purchased product | Binary indicator taking 1 if the product is purchased, 0 otherwise |
| Booking date | From 1 to 7, representing Monday through Sunday |
| Check-in date | From 1 to 7, representing Monday through Sunday |
| Check-out date | From 1 to 7, representing Monday through Sunday |
| Advanced purchase days | Time difference between the booking and check-in date |
| Party size | Number of adults and children associated with the booking |
| Length of stay | Number of nights (e.g. two) |
| Number of rooms | Number of rooms booked (e.g. three) |
| Membership status | Status in rewards program (0—not a member, 1—basic, 2—elevated, 3—premium) |
| Room type | Double, King, Queen, Regular, Special, Standard and Suite |
| Rate code | Rate1: Advance purchase |
|  | Rate2: Rack rate (Unrestricted rate) |
|  | Rate3: Rack rate combined with additional hotel service |
|  | Rate4: Accommodation with activities or awards |
|  | Rate5: Discounted rate less restricted than advance purchase |
| Price gap | Difference between the arrival price of one room product and the average price of the available choice set |

Table 2. Hotel case: feature list

| Clustering features ($s_{ij}^{ISF}$) | Choice model alternative specific features (part of of $s_{ij}^{MSF}$) | Choice model interactive features (here we use ~ to indicate interaction) (part of $s_{ij}^{MSF}$) |
| --- | --- | --- |
| Booking day of week | Price gap | Membership ~ price gap |
| Check-in day of week | Room type | Advance purchase days ~ price gap |
| Check-out day of week | Rate code | Income level ~ price gap |
| Advanced purchase days |  | Income level ~ room type (suite or not) |
| Party size |  | Income level ~ discount or not (based on rate code) |
| Length of stay |  | Length of stay ~ room type (suite or not) |
| Number of rooms |  | Length of stay ~ discount or not (based on rate code) |
| Membership |  | Party size ~ room type (suite or not) |
|  |  | Party size ~ discount or not (based on rate code) |

Table 7. Airline case: choice model feature list and description

| Features | Description |
| --- | --- |
| Fare gap | Difference between one itinerary's fare with the average in its choice set |
| First departure difference | Difference between one itinerary's first departure time and the individual's preferred first departure time |
| Second departure difference | Difference between one itinerary's second departure time and the individual's preferred second departure time |
| Length of stay ~ fare gap | Interactive terms between the trip's length of stay and the fare gap; length of stay is bucketed into three levels: level 1 (0~4 days), level 2 (4~7 days) and level 3 (7+ days) |
| Day of week ~ fare gap | Interactive terms between the trip's departure day of week and the fare gap; day of week is bucketed into three levels: early week (Monday, Tuesday and Wednesday), late week (Thursday and Friday) and weekend (Saturday and Sunday). |
| Advanced purchase days ~ fare gap | Interactive terms between individual's advanced purchase days and the fare gap; advanced purchase days is bucketed into four levels: early (84 days ahead), medium (28~84 days), late (14~28 days), rush (0~14 days). |
| Cabin ~ fare gap | Interactive terms between individual's preferred cabin with fare gap; based on fare class's seniority we categorized cabin into four levels: business, economy, discounted economy and no preference. |